%% file: main.tex
\documentclass[letterpaper, 10 pt, conference]{ieeeconf}  

\IEEEoverridecommandlockouts                              

\usepackage{mathptmx} 
\usepackage{times} 
\usepackage{bbding}
\usepackage{graphicx}
\usepackage{float}
\usepackage{diagbox}
\usepackage{arydshln}   
\usepackage{bbm}
\usepackage{wrapfig}
\usepackage{mathrsfs}
\usepackage{listings} 
\usepackage{physics}

\usepackage{pifont}  

\usepackage{graphics} 
\usepackage{float}
\usepackage{multirow}
\usepackage{amsmath}
\usepackage{bbm}
\usepackage{amssymb}
\usepackage{mathtools}
\usepackage[utf8]{inputenc} 
\usepackage[T1]{fontenc}    
\usepackage{hyperref}       
\usepackage{cleveref}
\usepackage{url}            
\usepackage{booktabs}       
\usepackage{amsfonts}       
\usepackage{nicefrac}       
\usepackage{microtype}      
\usepackage{xcolor}         
\usepackage{algorithm}
\usepackage{algorithmicx}
\usepackage{algorithm} 
\usepackage{algpseudocode}
\usepackage{fancyvrb}
\usepackage{fvextra}
\usepackage{csquotes}

\title{\LARGE \bf

RoboKeyGen: Robot Pose and Joint Angles Estimation via Diffusion-based 3D Keypoint Generation

}

\author{Yang Tian*, Jiyao Zhang*, Guowei Huang, Bin Wang, Ping Wang, Jiangmiao Pang, Hao Dong
\thanks{Yang Tian, Jiyao Zhang and Hao Dong are with CFCS, School of CS, Peking University and National Key Laboratory for Multimedia Information Processing.
Guowei Huang and Bin Wang are with Huawei.
Ping Wang is with School of Software \& Microelectronics and National Engineering Research Center for Software Engineering, Peking University.
Jiangmiao Pang is with the Chinese University of Hong Kong.
}
\thanks{* indicates equal contribution}
\thanks{
Corresponding to hao.dong@pku.edu.cn}
}

\begin{document}

\maketitle

\begin{abstract}
Estimating robot pose and joint angles is significant in advanced robotics, enabling applications like robot collaboration and online hand-eye calibration.
However, the introduction of unknown joint angles makes prediction more complex than simple robot pose estimation, due to its higher dimensionality.
Previous methods either regress 3D keypoints directly or utilise a render\&compare strategy. These approaches often falter in terms of performance or efficiency and grapple with the cross-camera gap problem.
This paper presents a novel framework that bifurcates the high-dimensional prediction task into two manageable subtasks: 2D keypoints detection and lifting 2D keypoints to 3D. 
This separation promises enhanced performance without sacrificing the efficiency innate to keypoint-based techniques.
A vital component of our method is the lifting of 2D keypoints to 3D keypoints. 
Common deterministic regression methods may falter when faced with uncertainties from 2D detection errors or self-occlusions.
Leveraging the robust modeling potential of diffusion models, we reframe this issue as a conditional 3D keypoints generation task.
To bolster cross-camera adaptability, we introduce the \emph{Normalised Camera Coordinate Space (NCCS)}, ensuring alignment of estimated 2D keypoints across varying camera intrinsics.
Experimental results demonstrate that the proposed method outperforms the state-of-the-art render\&compare method and achieves higher inference speed.
Furthermore, the tests accentuate our method's robust cross-camera generalisation capabilities.
We intend to release both the dataset and code in \href{https://nimolty.github.io/Robokeygen/}{https://nimolty.github.io/Robokeygen/}.
\end{abstract}

\vspace{-0.1cm}
\section{INTRODUCTION}
\vspace{-0.1cm}
Estimating robot pose and joint angles is crucial in intelligent robotics with implications for multi-robot collaboration~\cite{Rizk2019CooperativeHM}, online hand-eye calibration~\cite{Taunyazov2020EventDrivenVS}, and visual servoing~\cite{Chaumette2017ImageM} for close-loop control.
Extensive research has been conducted on robot pose estimation, such as marker-based easy-handeye~\cite{Tsai1988ANT} and learning-based online calibration methods~\cite{Lee2019CameratoRobotPE,Tian2023RobotSP,Lu2023MarkerlessCP}. 
However, these approaches assume known joint angles, a condition not always met. 
In multi-robot collaborations, for instance, state data may be unshared, necessitating concurrent robot pose and joint angle estimation.

\begin{figure}[t]
\begin{center}

    \includegraphics[width=0.8\linewidth]{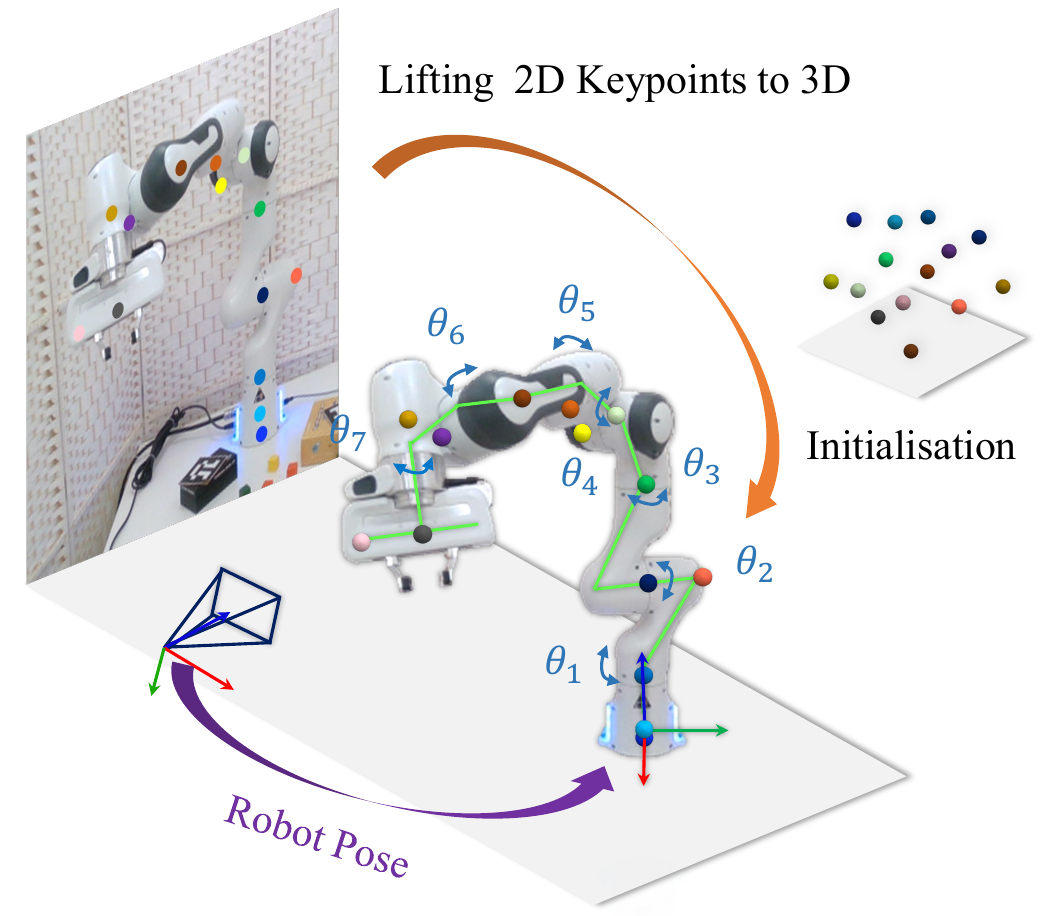}
\end{center}
\vspace{-0.3cm}
\caption{\textbf{RoboKeyGen.} Given RGB images, we aim to estimate the robot pose and joint angles. We achieve this goal by decoupling it into two more tractable tasks: 2D keypoints detection and lifting 2D keypoints to 3D.
}\label{fig:teasor}
\vspace{-0.7cm}
\end{figure}

Contrasting robot pose estimation with known versus unknown joint angles, the latter reveals heightened complexity due to increased degrees of freedom (\emph{e.g.}, from 6D to 13D for Franka). 
Existing methods can be divided into two categories: render\&compare approaches~\cite{Labbe2021SingleviewRP} and keypoints-based methods~\cite{Simoni2022SemiPerspectiveDH}.
RoboPose~\cite{Labbe2021SingleviewRP} extends render\&compare~\cite{li2018deepim, zakharov2019dpod} strategies from rigid object pose estimation~\cite{wang2019normalized, zhang2023genpose} to robot pose and joint angles estimation. 
However, this method suffers from slow inference speed (1 FPS in single frame mode) due to the iterative rendering. 
Conversely, SPDH~\cite{Simoni2022SemiPerspectiveDH} introduces a Semi-Perspective Decoupled Heatmaps representation, which extends the well-known 2D heatmaps to the 3D domain. 
It enables direct prediction of the 3D coordinates of predefined keypoints on the robot arm from a depth input and has a higher inference speed (22FPS) compared to render\&compare based methods. 
However, this approach exhibits a low accuracy and the proposed representation is theoretically limited by the presence of cross-camera generalisation issue.
In general, the existing methods are faced with such limitations:
\begin{itemize}
\item The conflict between efficiency and performance.
\item The cross-camera generalisation issue.
\end{itemize}

To address these challenges, we propose a novel framework named RoboKeyGen. 
The basic idea is illustrated in Fig.~\ref{fig:teasor}. 
Different from previous methods, we decouple this high-dimensional prediction task into two sub-tasks: 2D keypoints detection and lifting 2D keypoints to 3D. 
The former focuses on extracting the 2D keypoints from the appearance characteristics, while the latter concentrates on perspective transformation and the robot's structural information. 
This decoupling enables our method to improve performance while preserving the inherent efficiency of keypoints-based approaches.
Specifically, our method first predicts the 2D projections of predefined keypoints. 
Then, we align these projections into a normalised camera coordinate space.
Subsequently, we generate the 3D keypoints conditioned on the normalised 2D keypoints. These 3D keypoints are then utilised to regress joint angles. Finally, an off-the-shelf pose-fitting algorithm~\cite{Hua2020REDEEO} is employed to estimate the robot pose. 

Thanks to the significant advancements in 2D robot keypoints detection~\cite{Lee2019CameratoRobotPE}, we focus more on addressing the challenge of lifting these 2D keypoints to 3D and cross-camera generalisation. 
Direct regression proves suboptimal due to its failure to model the uncertainty brought by 2D keypoints detection errors. 
Instead, modeling the conditional distribution of 3D keypoints is more reasonable.
Leveraging the robust distribution modeling of diffusion-based models~\cite{Cai2020LearningGF,Wu2022TarGFLT,zhang2023genpose}, we employ a diffusion model conditioned on the estimated 2D keypoints to generate 3D keypoints. 
For cross-camera generalisation, considering the diverse camera intrinsic parameters has distinct projection transformations, we introduce the \emph{normalised camera coordinate space (NCCS)} for 2D keypoint alignment, effectively addressing the issue of cross-camera generalisation.

We provide a pipeline incorporating simulated training data and real-world datasets from two depth cameras for evaluation. 
Comparative analyses reveal our model's superiority over RoboPose~\cite{Labbe2021SingleviewRP} in performance and speed metrics, further underscoring its robustness in cross-camera generalisation.

\vspace{-0.1cm}
\section{RELATED WORKS}
\vspace{-0.1cm}

\subsection{Learning-based Robot Pose and Joint Angles Estimation}
\vspace{-0.1cm}
\subsubsection{Robot pose estimation with known joint angles}
Recent advances in deep learning offer innovative methods for robot pose recovery.
Dream~\cite{Lee2019CameratoRobotPE} uses a convolutional network to regress 2D heatmaps and compute poses through a \emph{Perspective-n-Point (PnP) RANSAC} solver~\cite{Lu2018ARO}. SGTAPose~\cite{Tian2023RobotSP} integrates temporal information to address self-occlusion in pose estimation. 
Meanwhile, CtRNet~\cite{Lu2023MarkerlessCP} employs a self-supervision framework, narrowing the sim-to-real gap effectively. Notably, these methods depend on immediate joint angles feedback, thus limiting their applicability.

\subsubsection{Robot pose and joint angles estimation}
When joint angles are unknown, methods fall into two main categories: render\&compare, and 3D keypoint detection.
RoboPose~\cite{Labbe2021SingleviewRP} offers a render\&compare framework for pose and joint angles using a single RGB image but is limited by a 1 FPS single-frame inference speed due to rendering. 
SPDH~\cite{Simoni2022SemiPerspectiveDH}, a depth-based approach, extends 2D to 3D heatmap pose estimation but faces a cross-camera challenge.
Our approach, in contrast, combines the speed of keypoint methods with a novel conditional 3D keypoints generation, addressing the cross-camera gap more effectively than SPDH~\cite{Simoni2022SemiPerspectiveDH}.

\vspace{-0.1cm}
\subsection{Diffusion Models}
\vspace{-0.1cm}
Diffusion models have gained significant attention in generative modeling. 
Some works have focused on theoretical aspects, such as training Noise Conditional Score Networks (SMLD) with denoising score matching objectives~\cite{Song2019GenerativeMB,Vincent2011ACB},
others introduced Denoising Diffusion Probabilistic Models (DDPM) that employ forward and reverse Markov chains~\cite{Ho2020DenoisingDP,SohlDickstein2015DeepUL}.
To provide a comprehensive understanding of these models, Song~\cite{Song2020ScoreBasedGM} presented a unified perspective that incorporates and explains the previously mentioned approaches.
Some studies also explored various applications of diffusion models, including medical imaging~\cite{Song2021SolvingIP},
point cloud generation~\cite{Cai2020LearningGF},
object rearrangement~\cite{Wu2022TarGFLT,wu2022example},
object pose estimation~\cite{Zhang2023GenPoseGC},
and human pose estimation~\cite{Ci2022GFPoseL3}.
Inspired by these advancements, we propose a novel diffusion-based framework for robot pose and joint angles estimation, specifically focusing on lifting 2D keypoints detection to conditional 3D keypoints generation.  
To the best of our knowledge, our method is the first exploration for learning the robot arm's structure via diffusion models.

\vspace{-0.1cm}
\section{METHOD}
\vspace{-0.1cm}
\begin{figure*}[t]
\begin{center}
    \includegraphics[width=\linewidth]{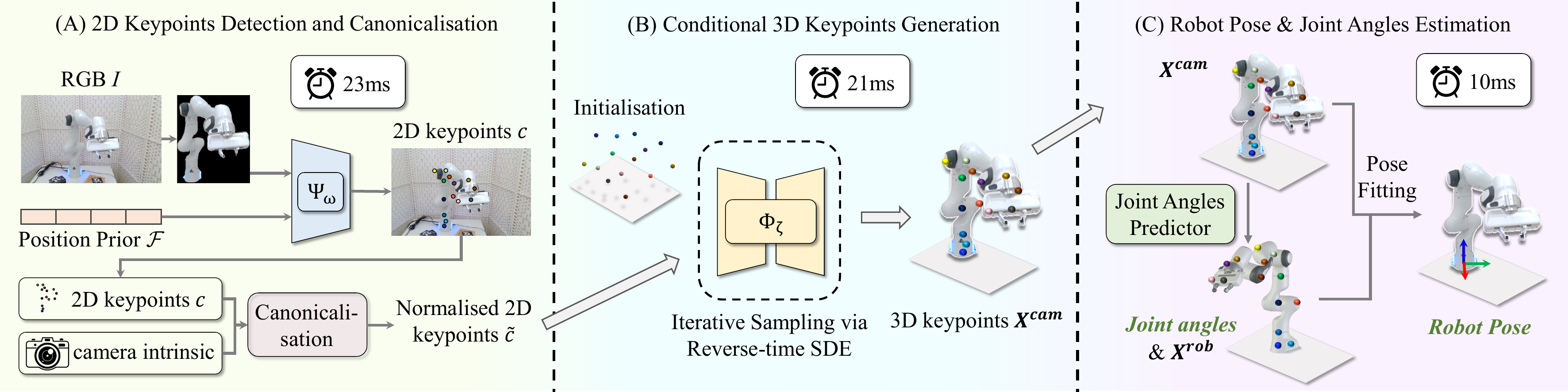}
\end{center}
\vspace{-0.3cm}
\caption{\textbf{The inference pipeline of RoboKeyGen.} (A) Combined with the RGB image $I$, predicted segmentation mask and positional embedding prior $\mathcal{F}$, we firstly predict 2D keypoints $c$ through the detection network $\Psi_{\omega}$. (B) Conditioning on 2D detections, we generate 3D $X^{cam}$ via the score network $\Phi_{\zeta}$. (C) Finally, we predict joint angles from $X^{cam}$ and recover $X^{rob}$ based on URDF files. We do pose fitting between $X^{cam}$ and $X^{rob}$ to acquire the robot pose.} \label{fig:Method}
\vspace{-0.6cm}
\end{figure*}

\noindent
\textbf{Task description}. Given a live stream of RGB images $\{I\}$, we aim to estimate the Robot Pose $ \{ \Gamma = (R, T) \in SE(3) \} $ and joint angles $ \{ \boldsymbol{\theta} \in \mathbb{R}^n \} $ (where n denotes the amounts of joint angles). Here we assume the forward kinematics and CAD models of the robot arm and camera intrinsics are known.

\noindent
\textbf{Overview}. We decouple the original high-dimensional task into two more tractable, low-dimensional sub-tasks: \textbf{2D keypoints detection} and \textbf{lifting 2D keypoints to 3D}. We first predict 2D projections of predefined keypoints $\boldsymbol{c}$ from RGB images $\boldsymbol{I}$. 
Then we align these estimated keypoints $\boldsymbol{c}$ into the form $\boldsymbol{\Tilde{c}}$ in \emph{Normalised Camera Coordinate Space (NCCS)}.
Further, a diffusion model $\Phi_{\zeta}$ is employed to model the distribution ($P_{data}(X^{cam} | \Tilde{c})$) of 3D keypoints $X^{cam}$ in camera space conditioned on normalised 2D keypoints $\boldsymbol{\Tilde{c}}$. Finally, we utilise a light regression network to predict joint angles $\boldsymbol{\theta}$ and recover 3D keypoints $X^{rob}$ in robot space. We restore the robot pose via pose fitting.

\vspace{-0.1cm}
\subsection{2D Keypoints Detection and Canonicalisation} \label{2D Keypoints Detection}
\vspace{-0.1cm}

We firstly detect 2D keypoints $c$ from RGB images. Then, considering that the distribution  of $X^{cam}$ conditioned on 2D keypoints projections $\boldsymbol{c}$ changes as camera intrinsics change,
we align $\boldsymbol{c}$ into normalised camera coordinate space $\boldsymbol{\Tilde{c}}$ to ensure a unique and  well-defined distribution $P_{data}(X^{cam} | \Tilde{c})$.

\subsubsection{2D Keypoints Detection}
We detect predefined 2D keypoints $c \in \mathbb{R}^{N\times2}$ from the current RGB frame $I$ and the last estimated 2D keypoints, where N denotes the amounts of keypoints. 
Specifically, to enable the 2D detection network $\Psi_{\omega}$ focus on extracting features from the pure robot arm and avoid disturbance from background textures, we first adopt the real-time semantic segmentation network PIDNet-L~\cite{Xu2022PIDNetAR} $M$ to segment the robot arm. 
Moreover, considering 2D keypoints between consecutive frames change slightly, we then project the estimated last frame's 2D projections into positional embedding priors $\mathcal{F}$ through sinusoidal transformations~\cite{Vaswani2017AttentionIA} and shallow MLPs as suggested in~\cite{Song2020ScoreBasedGM}.
Finally, given the RGB image $I$, segmentation mask and positional embedding $\mathcal{F}$ as input, an encoder-decoder detection network $\Psi_{\omega}$~\cite{Jiang2023RTMPoseRM, Li2021SimCCAS} is employed to predict the 2D keypoints $c$ of the current frame.
\subsubsection{2D Keypoints Canonicalisation}
For a given robot arm with available forward kinematics and predefined keypoints, we can easily find such an awkward property of $P_{data}(\boldsymbol{X^{cam}} | \boldsymbol{c})$:
For a common projection $\boldsymbol{c}$, cameras with different intrinsics yield diverse 3D Ground Truth (GT) keypoints, which makes the distribution $P_{data}(\boldsymbol{X} | \boldsymbol{c})$ poorly-defined.
To eliminate this issue, we project $\boldsymbol{c}$ into a normalised camera coordinate space (NCCS) $\boldsymbol{\Tilde{c}}$.
Specifically, with known camera intrinsics $\{f_x, f_y, c_x, c_y \}$, for $i-$th 2D keypoint $\boldsymbol{c^i}  = (u^i, v^i)\subset \boldsymbol{c}$, we transform $\boldsymbol{c^i}$ into $\boldsymbol{\Tilde{c}^i} = (\frac{u^i - c_x}{f_x}, \frac{v^i - c_y}{f_y})$.
According to the pinhole camera model (which is followed by most cameras in robotics), this transformation equals $\boldsymbol{\Tilde{c}^i} = (\frac{x^i}{z^i}, \frac{y^i}{z^i})$, where $(x^i, y^i, z^i) \subset \boldsymbol{X^{cam}}$ is the $i$-th keypoint's coordinates in camera space. 
Now we consider the new joint distribution $\Tilde{\mathcal{D}} = \{ (\Tilde{\boldsymbol{c}}, \boldsymbol{X^{cam}})=(\{ (\frac{x^i}{z^i}, \frac{y^i}{z^i}) \}_{i=1}^N, \{ (x^i, y^i, z^i) \}_{i=1}^N) \sim P_{data}(\Tilde{\boldsymbol{c}}, \boldsymbol{X^{cam}}) \}$.
We observe the condition $\Tilde{c}$ in $P_{data} (\boldsymbol{X^{cam}} | \boldsymbol{\Tilde{c}})$ is decoupled from camera intrinsics since it owns a normalised form regarding only coordinates in camera space.
In such situations, learning the new conditional distribution $P_{data} (\boldsymbol{X^{cam}} | \boldsymbol{\Tilde{c}})$ is essentially ensuring the z-coordinates for each keypoint.
In other words, our method only requires to concentrate on the robot arm's structure with no disturbance from camera intrinsics.

\vspace{-0.2cm}
\subsection{Conditional 3D Keypoints Generation via Diffusion Model}
\vspace{-0.1cm}
This section will illustrate how to sample the predefined 3D keypoints $\boldsymbol{X}^{cam}$ conditioned on the normalised 2D keypoints $\boldsymbol{\Tilde{c}}$ in a generative modeling paradigm.
Here we denote $\boldsymbol{X}\in \mathbb{R}^{N\times3}$ as 3D keypoints in camera space ($\boldsymbol{X}^{cam}$ in Fig. \ref{fig:Method}) for simplicity. 
We assume the 2D-3D keypoints pair in each image is sampled from an implicit joint distribution $\mathcal{D} = \{ (\boldsymbol{\Tilde{c}}, \boldsymbol{X}) \sim P_{data}(\boldsymbol{\Tilde{c}}, \boldsymbol{X}) \}$, and our objective is to model  $P_{data}(\boldsymbol{X} | \boldsymbol{\Tilde{c}})$.

\subsubsection{Learning the score function $\Phi_{\zeta}$}
We adopt a score-based diffusion model to model $P_{data} (\boldsymbol{X} | \boldsymbol{\Tilde{c}})$. 
Specifically, we take Variance Preserving (VP) Stochastic Differential Equation (SDE) proposed in~\cite{Song2020ScoreBasedGM} to construct a continuous time-dependent diffusion process $\{\boldsymbol{X}(t)\}_{t=0}^{T}$.
$\boldsymbol{X}(0)$ originates from $P_{data}(\boldsymbol{X} | \boldsymbol{c})$ and $\boldsymbol{X}(T)$ comes from the diffused prior distribution $p_T$.
As $t$ increases, $\{\boldsymbol{X}(t)\}_{t=0}^T$ is given by :
\begin{equation}
    d\boldsymbol{X} = -\frac{1}{2} \beta(t)\boldsymbol{X} dt + \sqrt{\beta(t)} d\boldsymbol{w} \label{VPSDE}
\end{equation}
where $\beta(t) = \beta(0) + t (\beta(1) - \beta(0))$. $\beta(0)$, $\beta(1)$ and $T$ are set as 0.1, 20.0 and 1.0 respectively. 

During Training, we aim to estimate the \emph{score function} of perturbed conditional distribution $\nabla_{\boldsymbol{X}} \log p_t(\boldsymbol{X} | \boldsymbol{\Tilde{c}})$ for all t:
\begin{equation} \label{pt}
    p_t(\boldsymbol{X}(t) | \boldsymbol{\Tilde{c}}) = \int p_{0t}(\boldsymbol{X}(t) | \boldsymbol{X}(0)) \cdot p_0 (\boldsymbol{X}(0) | \boldsymbol{\Tilde{c}}) d\boldsymbol{X}(0)
\end{equation}
where $p_{0t}$ is the transition kernel and $p_0 (\boldsymbol{X}(0) | \boldsymbol{\Tilde{c}})$ is exactly  $P_{data}(\boldsymbol{X} | \boldsymbol{\Tilde{c}})$.
$\nabla_{\boldsymbol{X}} \log p_t(\boldsymbol{X} | \boldsymbol{\Tilde{c}})$ can be estimated by training a score network $\boldsymbol{\Phi}_{\zeta}: \mathbb{R}^{3 \times N} \times \mathbb{R} \times \mathbb{R}^{2 \times N} \rightarrow{\mathbb{R}^{3 \times N}}$ via:
\begin{equation} \label{sde loss}
    \begin{aligned}
    \mathcal{L}(\zeta) &= \mathbb{E}_{t \sim \mathcal{U}(\epsilon, 1)} \{ \lambda(t) \mathbb{E}_{\boldsymbol{\Tilde{c}}, \boldsymbol{X}(0) \sim p_{data}(\boldsymbol{\Tilde{c}}, \boldsymbol{X})}  \mathbb{E}_{\boldsymbol{X}(t) \sim p_{0t}(\boldsymbol{X}(t)| \boldsymbol{X}(0))}\\
    & [ \Vert \boldsymbol{\Phi}_{\zeta} (\boldsymbol{X}(t), t | \boldsymbol{\Tilde{c}}) - \nabla_{\boldsymbol{X}(t)} \log p_{0t}(\boldsymbol{X}(t) | \boldsymbol{X}(0)) \Vert_{2}^2]  \} 
    \end{aligned}
\end{equation}
where $\epsilon$ is 0.0001 and  $\lambda (t)$ is set as $\beta (t)$ suggested in~\cite{Song2021MaximumLT}. 
The choice of VP SDE brings a closed form of $p_{0t}$ as follows:
\begin{equation}
    \begin{split}
\mathcal{N} (\boldsymbol{X}(t) ; \boldsymbol{X}(0) e^{-\frac{1}{2} \int_0^t \beta(s) ds}, \mathbf{I} - \mathbf{I} e^{-\int_0^t \beta(s) ds})
    \end{split}
\end{equation}
It is ensured that the optimal solution to Eq. \ref{sde loss}, denoted by $\boldsymbol{\Phi}_{\zeta^*} (\boldsymbol{X}, t | \Tilde{c})$ equals $\nabla_{\boldsymbol{X}} \log p_t (\boldsymbol{X} | \boldsymbol{\Tilde{c}})$ according to~\cite{Song2020ScoreBasedGM}.

\begin{table*}[tbp] 
\centering

\vspace{-0.1cm}
\caption{\textbf{Quantitative comparison with baselines.} \emph{Ours (single-frame)} and \emph{Ours (online)} denote initialization from Gaussian noise and the prediction of the last frame, respectively.
We also replace the backbone in~\cite{Simoni2022SemiPerspectiveDH} with resnet-101\cite{he2016deep} as another baseline \emph{SPDH-RESNET (Ours)}.
For a fair comparison, we train all the methods listed above on SimRGBD-Franka. 
} \label{main results}
\vspace{-0.1cm}

\input{tables/main_results}
\vspace{-0.7cm}
\end{table*}

\subsubsection{Sampling via the DDIM~\cite{Song2020DenoisingDI} sampler}
After training, we can sample $K$ groups of 3D Keypoints' candidates $\{ \boldsymbol{X_j}\}_{j=1}^{K}$ via diffusion samplers.

To speed up the inference phase, we select a fast DDIM sampler~\cite{Song2020DenoisingDI}. 
We iteratively generate $X(\tau_{t-1})$ from  $X(\tau_t)$ via the following equation:
\begin{equation} \label{ddim}
\begin{split}
    X(\tau_{t-1}) &= \sqrt{\overline{\alpha}_{\tau_{t-1}}} (\frac{X(\tau_{t})  - \sqrt{1 - \overline{\alpha}_{\tau_{t}}} \epsilon_{\zeta}(X(\tau_{t}), \tau_t | \Tilde{c})  }{\sqrt{\overline{\alpha}_{\tau_{t}}}}) \\
    &+ \sqrt{1 - \overline{\alpha}_{\tau_{t-1}} - \sigma_{\tau_t}^2  }  \cdot \epsilon_{\zeta}(X(\tau_{t}), \tau_t | \Tilde{c})  + \sigma_{\tau_t} \epsilon_{\tau_t} \\
\end{split}
\end{equation}
where $\{ \tau_i \}_{i=1}^{m}$ is the sampling timesteps. 

$\overline{\alpha}_{\tau_t}$, $\{ b_{\tau_i} \}_{i=1}^{m}$ and $\sigma_{\tau_t}$ remain the same notation and computation in~\cite{Ho2020DenoisingDP}.
$\epsilon_{\zeta}(X(\tau_{t}), \tau_t | \Tilde{c}) $ is the noise function and can be computed as $(- \sqrt{1 - e^{-\int_0^{\tau_{t}} \beta(s) ds}} \boldsymbol{\Phi}_{\zeta} (\boldsymbol{X}(\tau_{t}), \tau_{t} | \boldsymbol{\Tilde{c}}) )$.
In our implementation, we set $K$ as 10 and output the average value of  $\{\boldsymbol{X}(\tau_1)_j \}_{j=1}^K$.

\vspace{-0.2cm}
\subsection{Robot Pose and Joint Angles Estimation} \label{Robot Pose and Joint Angles Estimation}
\vspace{-0.1cm}
To further recover the robot's configuration, we target at estimating the robot's joint angles. 
Intuitively, we can connect the estimated 3D keypoints $\boldsymbol{X}^{cam}$ sequentially and regard them as a skeleton.
To estimate the joint angles from a skeleton, we only need to care about the positional relationship between adjacent "bones".
Hence, we train a simple MLP to directly regress joint angles $\boldsymbol{\theta}$ from $\boldsymbol{X}^{cam}$.
With available robot's forward kinematics and joint angles, we can recover the whole robot's configuration and compute $\boldsymbol{X}^{rob}$ according to the URDF file. 
Finally, we take a robust strategy using differentiable outliers estimation introduced in~\cite{Hua2020REDEEO} to implement the pose fitting between $\boldsymbol{X}^{cam}$ and $\boldsymbol{X}^{rob}$.

\vspace{-0.2cm}
\subsection{Implementation Details}
\vspace{-0.1cm}
To train the segmentation and detection network, we remain the same augmentations, loss functions and training strategies as suggested in~\cite{Xu2022PIDNetAR,Jiang2023RTMPoseRM}.
To train the score network $\Phi_{\zeta}$, we modify a vanilla fully connected network in~\cite{Ci2022GFPoseL3} as the backbone. 
We optimise the object in Eq. \ref{sde loss} for 2000 epochs with a batch size of 4096, learning rate 0.0002 and Adam optimiser.
To train the joint angle regression network, we design a shallow feedforward network. 
We train the network for 720 epochs with a batch size of 3600 via AdamW optimiser with initial learning rate 0.01 dropping by 0.1 at epoch 150, 300, 450.
See more details when code is released.

\vspace{-0.1cm}
\section{EXPERIMENTS AND RESULTS}
\vspace{-0.1cm}
\subsection{Datasets, Baselines and Metrics} 
\vspace{-0.1cm}
\subsubsection{Datasets} 
Since the public dataset in DREAM~\cite{Lee2019CameratoRobotPE} doesn't provide temporal images for training and lacks depth images, which are required for SPDH~\cite{Simoni2022SemiPerspectiveDH}, 
we propose three new datasets: a simulated training set, \textbf{SimRGBD-Franka},  and two real-world testing sets, \textbf{RealSense-Franka} and \textbf{AzureKinect-Franka} captured with different depth cameras.

\textbf{SimRGBD-Franka}: Following in~\cite{Tian2023RobotSP} and~\cite{Dai2022DomainRD}, we create this large-scale simulated dataset with Blender~\cite{Blender}. It comprises 4k videos, each with 3 consecutive frames, providing RGB images, robot pose, joint angles, masks, IR images, actual depth images, and simulated noisy depth images. 

\textbf{RealSense-Franka} and \textbf{AzureKinect-Franka}: Captured using external cameras (Realsense D415 and Microsoft Azure Kinect), these datasets showcase the Franka Emika Panda robot in motion. RealSense-Franka comprises 4 videos (3931 images), while AzureKinect-Franka has 5 videos (5576 images).
Each video starts with a stationary camera that eventually moves.
Regarding annotation, we firstly use COLMAP~\cite{colmap} to calibrate the camera extrinsics. 
Then, the initial frame in each video segment is manually annotated for robot pose and joint angles.
Finally, leveraging the calibrated camera extrinsics, we automatically get the annotations for the entire video segment. 
Both datasets include RGB images, robot pose, joint angles, and depth images.

\begin{figure*}[t]
\begin{center}
    \includegraphics[width=0.9\linewidth]{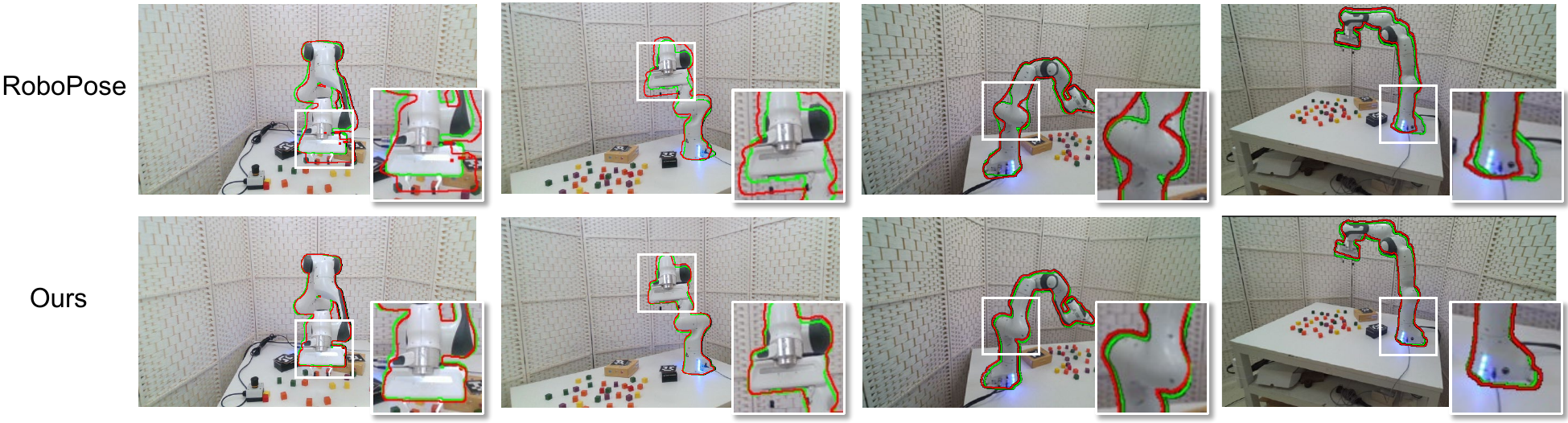}
\end{center}
\vspace{-0.5cm}
\caption{\textbf{Visualisation results on real-world datasets}. Green edges are ground truth while red edges are rendered via estimated robot pose and joint angles. White boxes highlight regions where ours  (online) performs better than RoboPose (online)~\cite{Labbe2021SingleviewRP}. 
}
\vspace{-0.4cm}
\label{fig:Visualize}
\end{figure*}

\subsubsection{Baselines} We compare our approach with previous methods in both unknown and known joint angles scenarios. 
\textbf{Unknown Joint Angles:}
\textbf{RoboPose}~\cite{Labbe2021SingleviewRP}: A state-of-the-art (SOTA) method that employs render\&compare to deduce joint angles and robot pose.
\textbf{SPDH}~\cite{Simoni2022SemiPerspectiveDH}: A direct method that derives 3D robot pose from a single depth map using semi-perspective decoupled heatmaps. 
\textbf{Known Joint Angles:}
\textbf{Dream}~\cite{Lee2019CameratoRobotPE}: An innovative technique that infers robot pose from a single frame via 2D heatmap regression and PnP-RANSAC solving.
\textbf{SGTAPose}~\cite{Tian2023RobotSP}: A pioneering approach that leverages temporal information for robot pose estimation.
\textbf{CtRNet}~\cite{Lu2023MarkerlessCP}: A pioneering approach that introduces a self-supervised strategy for online camera-to-robot calibration.

\subsubsection{Metrics}
We evaluate 3D metrics across all datasets. 
\textbf{ADD}: The average per-keypoint Euclidean norm between 3D keypoints and their transformed versions.
A lower ADD value reflects a higher pose estimation accuracy, improving downstream tasks (e.g. grasping) performances.
We compute the area under the curve (\textbf{AUC}) lower than a fixed threshold (10cm), median and mean values.
\vspace{-0.1cm}
\subsection{Comparison with Baselines}
\vspace{-0.1cm}
Table \ref{main results} showcases a notable performance enhancement of our method compared to state-of-the-art (SOTA) techniques.
In single-frame scenarios, our approach surpasses the current SOTA, RoboPose~\cite{Labbe2021SingleviewRP}, by 37.99\% and 28.72\% in AUC for RealSense-Franka and AzureKinect-Franka, respectively. Moreover, our inference speed increases from 1FPS to 12 FPS in the single-frame mode. This is due to the iterative rendering process involved in RoboPose, which is highly time-consuming.
In online scenarios, our method consistently outperforms RoboPose. 
Besides, the visualisation results in Figure \ref{fig:Visualize} also support our method's superiority, where white boxes highlight our better predictions than RoboPose's~\cite{Labbe2021SingleviewRP}. 
Our faster inference speed is attributed to the efficient DDIM sampler~\cite{Song2020DenoisingDI} and our online sampling strategy, both of which significantly reduce sampling steps.
Compared to the depth-based SPDH~\cite{Simoni2022SemiPerspectiveDH}, our method, despite a marginally slower inference speed, exhibits considerable advantages, particularly in AzureKinect-Franka. This is attributed to the theoretical limitations of the cross-camera generalisation issue. We will discuss this in Sec~\ref{Sec: cross-camera generalisation}
The commendable results of our approach can be credited to our decoupling scheme, allowing each module to specialise in a simpler sub-task.
\vspace{-0.1cm}
\subsection{Cross-Camera Generalisation Analysis}
\label{Sec: cross-camera generalisation}
\vspace{-0.1cm}
\begin{table}[htbp]
\centering

\vspace{-0.15cm}
\caption{Qualitative results of the cross-camera experiment. Results show that our method performs robustly across different cameras while SPDH~\cite{Simoni2022SemiPerspectiveDH} fluctuates dramatically.} \label{Cross Camera Tests}
\vspace{-0.2cm}
\input{tables/cross_camera}

\vspace{-0.4cm}
\end{table}
In practice, an ideal online calibration tool should adapt to different cameras. 
In this section, we evaluate our method's cross-camera generalisation capacity against the keypoint-based approach SPDH~\cite{Simoni2022SemiPerspectiveDH} in the context of unknown joint angles.
To ensure a fair comparison, we create three synthetic datasets emulating distinct camera fields of view (FOV): SimXBox360Kinect(FOV@62.73), SimRealSense(FOV@70.21), and SimAzureKinect(FOV@93.01), keeping other elements, such as robot pose, robot joint angles, and background, consistent.
Table \ref{Cross Camera Tests} reveals a notable performance decline for SPDH across varying cameras, while our method remains stable.
This observation is reinforced by results from the real datasets \textbf{RealSense-Franka} and \textbf{AzureKinect-Franka} in Table \ref{main results}.
The primary reason for this substantial difference is that SPDH relies on XYZ-maps as inputs and employs convolutional networks as backbones. 
Consequently, when the topological structures of XYZ-maps are transformed due to changes in camera intrinsics, the translation invariance property of convolutional networks leads to misguided predictions in SPDH's UZ map.
Our method's resilience is attributed to our task decoupling.
Specifically, for the conditional 3D keypoints generation, we employ normalised camera coordinates $\boldsymbol{\Tilde{c}}$, unaffected by camera intrinsic alterations. 
Additionally, prior studies~\cite{Lee2019CameratoRobotPE} have already proved the satisfying cross-camera generalisation capacity of 2D keypoints detection.

\vspace{-0.1cm}
\subsection{Ablaton Studies}
\vspace{-0.1cm}
\subsubsection{Conditional generation vs. regression}
\begin{table}[htbp]
\centering
\vspace{-0.2cm}
\caption{
Ablation between generation and regression. 
}\label{regression}
\vspace{-0.3cm}
\input{tables/regression}
\vspace{-0.5cm}
\end{table}
Here we evaluate the efficacy of our conditional 3D generation module against direct regression. 
Utilising the framework by Martinez et al.~\cite{martinez2017simple}, we adapt it as a regression baseline to transform 2D keypoints into 3D.
Table \ref{regression} contrasts our method with this baseline, underscoring a significant enhancement with our method.
This improvement can be credited to the generative models' advanced nonlinear modeling capabilities and their resilience to noise disturbances frequently observed in 2D keypoints detection, such as missing or noisy keypoints.

\subsubsection{Normalised camera coordinate space (NCCS)}
\vspace{-0.2cm}
\begin{table}[htbp]
\centering
\caption{Importance of conditioning on nccs.}\label{nccs_test}
\vspace{-0.2cm}
\input{tables/nccs_test}

\vspace{-0.35cm}
\end{table}

Here we highlight the impact of normalised camera coordinates space. 
In Table \ref{nccs_test}, w/o NCCS indicates models trained solely on raw 2D keypoints. 
Notably, w/o NCCS exhibits a substantial error, approximating 40cm in both median and mean ADD.
The reason behind this exceptionally poor performance is that the diffusion model not only needs to memorise the robot's structural information, but also excessively fits to the fixed training camera intrinsics and projection formula. 
Therefore, when exposed to a novel camera,  \emph{w/o NCCS} struggles to adjust to the altered intrinsics. 
Conversely, the integration of NCCS provides a standardised 2D representation, effectively mitigating disruptions from diverse intrinsics and ensuring consistent performance across varying cameras.

\subsubsection{Samplers and initializations}
\vspace{-0.1cm}
\begin{table}[htbp]
\centering
\vspace{-0.2cm}
\caption{Ablation on different samplers and initializations.
} \label{tracking_single}
\vspace{-0.2cm}

\input{tables/tracking_single}
\vspace{-0.4cm}
\end{table}
We investigate the impact of different sampling solvers, specifically ODE~\cite{Dormand1980AFO} and DDIM~\cite{Song2020DenoisingDI}, coupled with distinct initialization techniques on the sampling procedure. Table~\ref{tracking_single} demonstrates that the $\emph{DDIM}$ solver significantly reduces sampling time compared to the $\emph{ODE}$ solver, yet maintains comparable performance. Furthermore, $\emph{Online}$ initialization consistently outperforms the initialization from Gaussian noise in terms of both inference speed and performance, regardless of the type of solvers.
This superiority of the $\emph{Online}$ initialisation can be attributed to its use of predictions from the last frame, which offers an initialisation proximate to the genuine distribution. 
Consequently, this enables shorter sampling steps and helps circumvent certain local optima. All the inference speeds were tested using a single V100 GPU.

\subsubsection{Number of candidates}

\begin{figure}[t]
\begin{center}
    \includegraphics[width=0.9\linewidth]{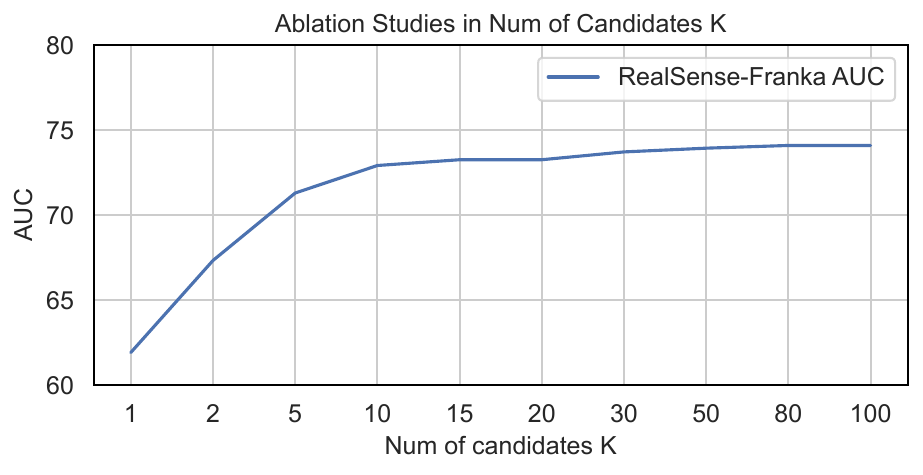}
\end{center}
\vspace{-0.5cm}
\caption{Ablation on different number of 3D keypoints candidates $K$. We finally adopt $K=10$ in implementation.}\label{fig:Num of candidates}
\vspace{-0.7cm}
\end{figure}

\begin{table}[htbp]
\centering
\caption{Additional comparison with baselines in settings with known joint angles. "-" denotes errors larger than 5m.}\label{gt_angles}
\vspace{-0.2cm}

\input{tables/gt_angles} 
\vspace{-0.7cm}
\end{table}

Figure \ref{fig:Num of candidates} elucidates the impact of the number of 3D keypoints' candidates $K$ during inference time. 
Regarding the AUC in RealSense Franka, the network's performance shows a great enhancement when $K$ rises from 1 to 10.
This can be explained that the augmented size of samples leads to a keypoints candidate set more closely aligned with the predicted distribution.
Nonetheless, the enhancement becomes marginal as $K$ extends to 100, 
likely due to the mean predictions nearing the upper limit of the sampling strategy.
In view of the trade-off between performance and overhead, we adopt $K=10$.

\vspace{-0.1cm}
\subsection{Additional comparison in settings with known joint angles.}
\vspace{-0.1cm}
While our primary emphasis is on estimating the robot pose and joint angles, our method demonstrates a marked advantage over prior methods with known joint angles.
We employ ground truth joint angles to reconstruct $\boldsymbol{X}^{rob}$ in Sec \ref{Robot Pose and Joint Angles Estimation}, devoid of any specific additional design.
As illustrated in Table \ref{gt_angles}, our method consistently outperforms others across all evaluation metrics. 
Notably, relative to SGTAPose~\cite{Tian2023RobotSP}, which integrates temporal information, our method exhibits a 20\% enhancement in AUC.
Moreover, against the CtRNet~\cite{Lu2023MarkerlessCP} that relies on additional real images for self-supervision, our method maintains superior performance with respect to ADD median and mean, achieving an average decrease of nearly 1 cm and 2.5 cm, respectively.
Interestingly, our method doesn't improve much with known joint angles, and we attribute this to the robust pose fitting strategy in \cite{Hua2020REDEEO}. 

\vspace{-0.15cm}
\section{CONCLUSION AND DISCUSSION}
\vspace{-0.1cm}
In this paper, we tackle the challenges in robot pose and joint angles estimation, specifically the efficiency-performance trade-off and cross-camera generalisation. 
To this end, we propose a novel framework named RoboKeyGen, 
which decouples this task into 2D keypoints detection and lifting 2D keypoints to 3D.
Our method achieves high performance while preserving the efficiency inherent in keypoints-based methods.
Our diffusion-based, conditional 3D keypoints generation effectively manages uncertainties arising from errors in 2D keypoints detection.  
Moreover, incorporating \emph{Normalised Camera Coordinate Space} (NCCS) handles cross-camera generalisation issue. 
Experimental results show the effectiveness of our approach over state-of-the-art methods. 
\textbf{Limitations and future works}: Although our method outperforms render\&compare based methods in performance and inference speed (18 FPS), it doesn't meet real-time requirements in  certain scenarios. Moreover, 
we didn't take the scenes where robots are partially occluded or truncated into consideration.
Future research could explore algorithms with a faster inference speed and robust to occlusion.

\vspace{-0.15cm}
\section{Acknowledgement}
\vspace{-0.1cm}
This work is supported by the National Youth Talent Support Program (Project ID: 8200800081), and National Natural Science Foundation of China (Project ID: 62136001).
%

\clearpage

{
\bibliographystyle{IEEEtran}
\bibliography{IEEEabrv,reference}
}

\end{document}

%% file: tables/main_results.tex
\begin{tabular}{cccccccc}
\toprule
\multirow{2}{*}{Method} & \multicolumn{3}{c}{RealSense-Franka} & \multicolumn{3}{c}{AzureKinect-Franka} & \multirow{2}{*}{FPS} \\ \cmidrule{2-7} 
 & AUC@0.1m$\uparrow$ & Median(m)$\downarrow$ & Mean(m)$\downarrow$ & AUC@0.1m$\uparrow$ & Median(m)$\downarrow$ & Mean(m)$\downarrow$ & \\ \midrule \midrule
RoboPose (single-frame)~\cite{Labbe2021SingleviewRP} & 29.01          & 0.081           & 0.116          & 32.00       & 0.069          & 0.106     & 1       \\  
RoboPose (online)~\cite{Labbe2021SingleviewRP} & 31.78          & 0.073           & 0.105          & 39.83       & 0.047          & 0.083    & 16         \\  
SPDH-HRNet~\cite{Simoni2022SemiPerspectiveDH}           & 3.39            & 1.366         & 1.297         & 0.00          & 0.643          & 0.669     & 10       \\ 
SPDH-SH~\cite{Simoni2022SemiPerspectiveDH}      & 17.24            & 0.251          & 0.272          & 0.00            & 0.844          & 0.854     & \textbf{22}       \\
SPDH-RESNET~\cite{Simoni2022SemiPerspectiveDH} (Ours)          & 19.46            & 0.090          & 0.135          & 0.00            & 0.793          & 0.789  & 18          \\ 
Ours (single-frame) & 67.00 & 0.027 & 0.035 & 60.72 & 0.030 & 0.049 
& 12 \\
Ours (online)   & \textbf{72.93}  & \textbf{0.022}  & \textbf{0.028}  & \textbf{63.33}  & \textbf{0.028}  & \textbf{0.045}   & 18          \\ \bottomrule
\end{tabular}

%% file: tables/cross_camera.tex
\begin{tabular}{cccc}
\toprule
Method     &AUC@0.1m  & Median(m) & Mean(m)  \\ \midrule \midrule
  \multicolumn{4}{c}{SimXBox360Kinect (FOV@62.73)} \\ \midrule
SPDH-HRNET~\cite{Simoni2022SemiPerspectiveDH}    &   60.80      & 0.034       & 0.058   \\
SPDH-SH~\cite{Simoni2022SemiPerspectiveDH}       &   71.20      & 0.027       & 0.029   \\
SPDH-RESNET~\cite{Simoni2022SemiPerspectiveDH} (Ours)       &   69.84      & 0.029       & 0.030   \\
Ours (online) & \textbf{77.47} & \textbf{0.017}      & \textbf{0.025}     \\ \midrule
 \multicolumn{4}{c}{SimRealSense (FOV@70.21)} \\ \midrule
SPDH-HRNET~\cite{Simoni2022SemiPerspectiveDH}    &   7.88     & 1.269     & 1.084 \\
SPDH-SH~\cite{Simoni2022SemiPerspectiveDH}       &  61.37     & 0.038       & 0.039    \\
SPDH-RESNET~\cite{Simoni2022SemiPerspectiveDH} (Ours)       &   58.98     & 0.040       & 0.041   \\
Ours (online) & \textbf{76.53} & \textbf{0.018}       & \textbf{0.026}    \\ \midrule
 \multicolumn{4}{c}{SimAzureKinect (FOV@93.01)} \\ \midrule
SPDH-HRNET~\cite{Simoni2022SemiPerspectiveDH}      &  0.00    & 2.097     & 2.064  \\
SPDH-SH~\cite{Simoni2022SemiPerspectiveDH}        &   0.04    & 0.413      & 0.449   \\
SPDH-RESNET~\cite{Simoni2022SemiPerspectiveDH} (Ours)       &   0.00     & 0.211       & 0.233   \\
Ours (online) & \textbf{69.01} & \textbf{0.022}     & \textbf{0.040}  \\ \bottomrule
\end{tabular}

%% file: tables/regression.tex

\begin{tabular}{ccccc}
\toprule
Method     &  2D &     AUC@0.1m & Median(m) & Mean(m) \\ \midrule \midrule
             \multicolumn{5}{c}{RealSense-Franka}             \\ \midrule
Regression & GT  &  37.48  & 0.060       & 0.067    \\
Regression & Prediction &  30.84  & 0.064       & 0.084    \\
Generation & GT & \textbf{83.51} & \textbf{0.015} & \textbf{0.016} \\
Generation & Prediction & 72.93  & 0.022       & 0.028    \\ \midrule
             \multicolumn{5}{c}{AzureKinect-Franka}         \\ \midrule
Regression & GT  &  43.28  & 0.049       & 0.067    \\
Regression & Prediction &  37.66  & 0.058       & 0.076    \\
Generation & GT & \textbf{82.33} & \textbf{0.016} & \textbf{0.018} \\
Generation & Prediction & 63.33  & 0.028       & 0.046    \\ \bottomrule
\end{tabular}


%% file: tables/nccs_test.tex

\begin{tabular}{cccc}
\toprule
Method               &       AUC@0.1m & Median(m) & Mean(m) \\ \midrule \midrule
               \multicolumn{4}{c}{RealSense-Franka}             \\ \midrule
w/o NCCS (online) &  0.00  & 0.413       & 0.434    \\
Ours (online) &  \textbf{72.93}  & \textbf{0.022}       & \textbf{0.028}    \\ \midrule
                \multicolumn{4}{c}{AzureKinect-Franka}         \\ \midrule
w/o NCCS (online) &  0.00  & 0.397       & 0.412    \\
Ours (online)  &  \textbf{63.33}  & \textbf{0.028}       & \textbf{0.046}    \\ \bottomrule
\end{tabular}


%% file: tables/tracking_single.tex
\begin{tabular}{cccccc}
\toprule
Sampler   & Online  &      AUC@0.1m & Median(m) & Mean(m)    &   FPS\\ \midrule \midrule
              \multicolumn{6}{c}{RealSense-Franka}             \\ \midrule
ODE    &  \ding{55} & 64.85  &    0.032  & 0.037    & 1      \\
DDIM   &  \ding{55} & 67.00  &  0.027    & 0.035    & 12     \\
ODE    &  \checkmark & \textbf{73.94}  &  \textbf{0.021}   & \textbf{0.027}    & 14    \\ 
DDIM   &  \checkmark & 72.93  &    0.022  & 0.028    & \textbf{18}    \\ \midrule

              \multicolumn{6}{c}{AzureKinect-Franka}         \\ \midrule
ODE    & \ding{55} &  62.61 &    0.029  & 0.047   & 1     \\
DDIM   & \ding{55} &  60.72  &   0.030   & 0.049   & 12     \\
ODE    & \checkmark &  62.90  &    0.029  & \textbf{0.044}   & 14   \\ 
DDIM   & \checkmark &  \textbf{63.33}   &    \textbf{0.028}  & 0.045   & \textbf{18}     \\ \bottomrule

\end{tabular}

%% file: tables/gt_angles.tex
\begin{tabular}{cccc}
\toprule
Method               &       AUC@0.1m & Median(m) & Mean(m) \\ \midrule \midrule
               \multicolumn{4}{c}{RealSense-Franka}             \\ \midrule
Dream-VGG-Q~\cite{Lee2019CameratoRobotPE}  &  27.48          & 0.080          & 0.244  \\ 
Dream-VGG-F~\cite{Lee2019CameratoRobotPE}  & 2.31           & 1.385          & - \\
Dream-RESNET-H~\cite{Lee2019CameratoRobotPE} &       40.75     &      0.053     &     0.177  \\
Dream-RESNET-F~\cite{Lee2019CameratoRobotPE}  & 20.31           & 1.095          & - \\
RoboPose (single-frame)~\cite{Labbe2021SingleviewRP} & 44.21          & 0.050           & 0.062 \\
RoboPose (online)~\cite{Labbe2021SingleviewRP} & 44.18          & 0.050           & 0.062 \\
SGTAPose~\cite{Tian2023RobotSP} &      52.00      &     0.036     &       1.370 \\
CtRNet~\cite{Lu2023MarkerlessCP}&      59.51           &        0.031         &        0.056 \\
Ours (single-frame) & 68.34  & 0.025  & 0.033 \\
Ours (Online) & \textbf{74.76}  & \textbf{0.020}  & \textbf{0.026} \\ \midrule
                \multicolumn{4}{c}{AzureKinect-Franka}         \\ \midrule
Dream-VGG-Q~\cite{Lee2019CameratoRobotPE}  & 32.35          & 0.075           & 0.352 \\ 
Dream-VGG-F~\cite{Lee2019CameratoRobotPE} &  0.37              & 1.471          & - \\
Dream-RESNET-H~\cite{Lee2019CameratoRobotPE} &   51.05         &      0.038      &    0.133  \\
Dream-RESNET-F~\cite{Lee2019CameratoRobotPE}  & 38.60          & 0.053           & - \\
RoboPose (single-frame)~\cite{Labbe2021SingleviewRP} & 41.50          & 0.053           & 0.062 \\
RoboPose (online)~\cite{Labbe2021SingleviewRP} & 41.59         & 0.053           & 0.062 \\
SGTAPose~\cite{Tian2023RobotSP} &       44.80     &     0.050      &     0.129 \\
CtRNet~\cite{Lu2023MarkerlessCP}&    55.22             &        0.035         &      0.062   \\
Ours (single-frame)& 63.74  & 0.027  & 0.045 \\
Ours (Online) & \textbf{66.84}  & \textbf{0.025}  & \textbf{0.041}  \\ \bottomrule
\end{tabular}